\documentclass[10pt,journal,compsoc]{IEEEtran}

\ifCLASSOPTIONcompsoc
  \usepackage[nocompress]{cite}
\else
  \usepackage{cite}
\fi
\ifCLASSINFOpdf
\else
\fi
\usepackage{amsmath}

\DeclareMathOperator*{\argmin}{arg\,min}

\usepackage{hyperref}

\usepackage{amssymb}
\usepackage{pifont}
\newcommand{\cmark}{\ding{51}}%
\newcommand{\xmark}{\ding{55}}%
\usepackage{booktabs}
\usepackage{makecell}
\usepackage{pbox}
\usepackage{xcolor}
\usepackage{adjustbox}

\usepackage{caption} 
\usepackage{graphicx}
\usepackage{subcaption}

\usepackage{color, colortbl}
\definecolor{Gray}{gray}{0.9}

\usepackage{svg}

\usepackage{pgfplots}
\usepgfplotslibrary{groupplots}
\DeclareUnicodeCharacter{2212}{−}
\usepgfplotslibrary{groupplots,dateplot}
\usetikzlibrary{patterns,shapes.arrows}
\pgfplotsset{compat=newest}

\begin{document}
%
\title{Multimodal Engagement Analysis from Facial Videos in the Classroom}
%
%
%
%


\author{Ömer~Sümer,
        Patricia~Goldberg,
        Sidney D'Mello,
        Peter~Gerjets,
        Ulrich~Trautwein,
        and~Enkelejda~Kasneci 
\IEEEcompsocitemizethanks{
\IEEEcompsocthanksitem Ö. Sümer, P. Goldberg, and U. Trautwein are with the Hector Research Institute of Education Sciences and Psychology, University of Tübingen, Tübingen, 72072, Germany, Ö. Sümer is also with the Department of Computer Science, University of Tübingen, Tübingen, 72076, Germany.
\IEEEcompsocthanksitem Sidney K. D'Mello is with the Institute of Cognitive Science and the Department of Computer Science, University of Colorado Boulder, Boulder,
CO 80309. E-mail: sidney.dmello@colorado.edu
\IEEEcompsocthanksitem P. Gerjets is with the Leibniz-Institut für Wissensmedien, Tübingen, 72076, Germany. E-mail: p.gerjets@iwm-tuebingen.de.
\IEEEcompsocthanksitem E. Kasneci is with the Department of Computer Science, University of Tübingen, Tübingen, 72076, Germany. E-mail: enkelejda.kasneci@uni-tuebingen.de.\protect\\
}
}

%
%

\markboth{}%
{Shell \MakeLowercase{\textit{et al.}}: Bare Demo of IEEEtran.cls for Computer Society Journals}
%




\IEEEtitleabstractindextext{%
\begin{abstract}
Student engagement is a key construct for learning and teaching. While most of the literature explored the student engagement analysis on computer-based settings, this paper extends that focus to classroom instruction. To best examine student visual engagement in the classroom, we conducted a study utilizing the audiovisual recordings of classes at a secondary school over one and a half month's time, acquired continuous engagement labeling per student (N=15) in repeated sessions, and explored computer vision methods to classify engagement levels from faces in the classroom. We trained deep embeddings for attentional and emotional features, training Attention-Net for head pose estimation and Affect-Net for facial expression recognition. We additionally trained different engagement classifiers, consisting of Support Vector Machines, Random Forest, Multilayer Perceptron, and Long Short-Term Memory, for both features. The best performing engagement classifiers achieved AUCs of .620 and .720 in Grades 8 and 12, respectively. We further investigated fusion strategies and found score-level fusion either improves the engagement classifiers or is on par with the best performing modality. We also investigated the effect of personalization and found that using only 60-seconds of person-specific data selected by margin uncertainty of the base classifier yielded an average AUC improvement of .084.Our main aim with this work is to provide the technical means to facilitate the manual data analysis of classroom videos in research on teaching quality and in the context of teacher training. 
\end{abstract}

}

\maketitle

%

\IEEEpeerreviewmaketitle

\IEEEraisesectionheading{\section{Introduction}\label{section:introduction}}

\IEEEPARstart{W}{hich}  students are engaged in learning during the class?  What is the relationship between student engagement and the content and the quality of the learning material? And, additionally, how can we relate student engagement to learning outcomes or long-term goals? These research questions and more drew the interest of scientists from educational sciences, psychology, and similar fields to investigate student engagement.

To begin our investigation of student engagement, we must first define the term engagement and contextualize its implications in the classroom setting. Being engaged means ``to involve oneself or become occupied; to participate'' while engagement can be defined as ``[being] actively committed''. As it relates to human behavior, engagement is highly connected to commitment and involvement. In the educational context, according to Fredricks et al.'s widely accepted definition \cite{Fredricks:2004}, engagement is a multidimensional construct that is composed of three dimensions: \emph{behavioural}, \emph{cognitive}, and \emph{emotional}. Those dimensions do not reflect isolated processes, but rather dynamically interrelated factors within an individual student. Behavioral engagement focuses on the act of participation and can include behaviors such as displaying attention and concentration, or asking questions. The basis of behavioral engagement is involvement in social, academic, or extracurricular activities seen as essential for achieving positive outcomes. On the other hand, emotional engagement encompasses affective components such as students' interest or boredom. Whereas aspects of behavioral and emotional engagement are typically observable from the outside, cognitive engagement incorporates less overt internal, cognitive processes such as psychological resource investments in learning and self-regulation \cite{Fredricks:2004}. It also incorporates the students' willingness to comprehend complex ideas and master skills. Importantly, previous research found positive correlations between aspects of student engagement and academic achievement, emphasizing student engagement's central role in classroom learning \cite{Lei:2018}. 

In the present study, we aim to evaluate student engagement based on visible indicators in learning situations and approach the analysis from an affective computing perspective. Two methods are proposed by affective computing and classroom management literature to acquire engagement levels: 1) self-reports and 2) observer ratings. Self-reports are practical, relatively cheap, and easy to administer to a large sample, making them valuable in various tasks related to engagement and beyond \cite{Christenson:2012:handbook}. Despite their value, self-reports have certain drawbacks, namely a dependence on participant compliance, diligence, and a student's overall understanding of being engaged. These characteristics, however, are not always a given, especially with younger students.

External behavior observations are another useful assessment tool for student engagement as they have a long tradition in education research to investigate determinants of classroom processes, such as the quality of instruction~\cite{van2017measuring}, teacher-student relationships (e.g., \cite{pianta2008classroom}), amount of learning opportunities (e.g.,~\cite{karweit1981measurement}), or teachers’ choice of practices (e.g.,~\cite{beyda2002relationship}). In general, observer ratings are systematic approaches that aim to detect and interpret certain behaviors \cite{Girard:2016}. Their deployment in large-scale studies is notably limited by the necessity of providing human raters with specialized training and the difficulty of acquiring reliable labeling. Moreover, in contrast to many other computer vision applications, crowdsourcing is not a viable option to label student engagement due to ethical considerations and the specialized training required for the raters.

Owing to the limitations of self-reports and observer ratings, automated approaches for estimating student engagement pose a challenge when increasing the sample size of classroom observation studies. A solution is to automatically estimate engagement using machine learning and computer vision. In the field of affective computing, initial studies aimed at estimating student engagement focused on computer-based learning and intelligent tutor systems (ITS). From ITS log files such as students' reaction times,  errors, and performance, preferred modalities for engagement analyses shifted to video, audio, and physiological measures (i.e., galvanic skin response, EEG, heart rate).

In computer-based learning settings, the availability of log data is an important asset. Furthermore, vision-based features can be extracted reliably using webcams. In the classroom, on the contrary, using sensors for each student can render studies expensive and intrusive and ultimately may affect student behaviors. Thus, a widely accepted practice in classrooms is to record the instruction with field cameras in the corners of the room. One drawback of this approach, however, is that the audio and visual data is noisy and may be occluded. 

\subsection{Contributions of the Study}
In this study, we review, in detail, engagement studies in the field of affective computing. We then discuss the large-scale school study we conducted by collecting audio-visual recordings of classes during a one and half month period. Observer ratings of student engagement were acquired using an instrument previously validated in university-level seminars \cite{Goldberg:2019}. 

The current study's primary focus is to learn engagement classification from limited and unconstrained data where traditional face alignment and facial action unit estimation methods have largely failed. Following the definition by Fredricks et al. \cite{Fredricks:2004}, behavioral and emotional aspects of student engagement can best be observed from the outside. Visual attention (subsequently referred to as attention) and affective components can thus serve as approximations of these two sub-dimensions. We propose learning attention and affect features from two convolutional neural networks trained on head pose estimation and facial expression recognition as pretask. In contrast to previous works that utilize faces based on handcrafted features in engagement analysis, the deep learning-based representations we propose work without precise facial alignment.

Our engagement classification is performed in these learned feature embeddings. We also applied feature and score level fusion on attention and affect features. Beyond the person-independent evaluation training and evaluation of engagement classifiers, we also investigated personalization because there is intrapersonal variation in students' (dis)engagement. 

Although automated engagement analysis is widely studied in computer-based settings such as intelligent tutors and educational games, to our knowledge this study is one of the first to perform video-based engagement classification in the classroom on a large scale.

\begin{table*}[ht]
\caption{Automated Engagement Analysis in Classroom, Computer-based Learning, Human-Human/Human-Robot Intreraction (HHI/HRI) Settings}
\centering
\begin{adjustbox}{max width=\textwidth}
\begin{tabular}{llllc}
\toprule 
\textbf{Reference}  & \textbf{Setting} & \textbf{Behavioral Cues} & \textbf{Engagement Measurement} &  \textbf{Predictive Models}  \\ \toprule
\cite{Bidwell:2011}  & classroom & head pose & observer reports & \cmark \\ \hline
\cite{Raca:2013:a, Raca:2015:dissertation}  & classroom & head pose, body motion & self-reports (in-class) &  \xmark \\  \hline
\cite{Zaletelj:2017:a, Zaletelj:2017:b}  & classroom & head pose, gaze, facial expressions, posture & observer reports & \cmark \\ \hline
\cite{Fujii:2018}  & classroom & gaze mapping (heads up/down) & -- & \xmark \\ \hline
\cite{Thomas:2017} & classroom & head pose, gaze, FACS action units & observer reports & \cmark \\ \hline
\cite{Ahuja:2019}  & classroom & \pbox{15cm}{real-time monitoring system capable of extracting \\ many behavioral features (i.e. smile detector,\\ hand raising, head pose, speech analysis)} & -- & \xmark \\ \hline
\cite{Anh:2019}  & classroom & \pbox{15cm}{monitoring system \\ (head pose and gaze estimation)} & -- & \cmark \\  \toprule
\cite{D'Mello:2009} & computer-based & FACS action units and ITS log features & observer ratings &  \cmark \\ \hline
\cite{Grafsgaard:2012} & computer-based & FACS action units & \pbox{15cm}{self-reports \\ (user engagement survey \cite{O'Brien:2010},\\ NASA-TLX \cite{Hart:1988})} & \cmark \\ \hline
\cite{Whitehill:2014}  & computer-based & handcrafted features from faces & observer reports & \cmark \\ \hline
\cite{Bosch:2016:a}  & computer-based & FACS action units and appearance features & self-/observer reports (MW) & \cmark \\ \hline
\cite{Bosch:2016:b}  & computer-based & FACS action units and gross body movement & observer reports (BROMP \cite{Ocumpaugh:2015})  & \cmark \\ \hline
\cite{Monkaresi:2017}  & computer-based & \pbox{15cm}{Kinect Animation Units, facial appearance, \\ heart rate estimated from face videos} & \pbox{15cm}{self-reports \\ (concurrent \& retrospective)} & \cmark \\ \hline
\cite{Kamath:2016} & computer-based & facial appearance features & crowdsourcing & \cmark  \\ \hline
\cite{Kaur:2018}  & computer-based & head pose and gaze direction & observer reports & \cmark  \\ \toprule 
ELEA \cite{Sanchez-Cortes:2011} & HHI & --  & observer ratings &  \xmark \\ \hline
RECOLA \cite{Ringeval:2013}  & HHI & --  & self-reports & \xmark \\ \hline
MHHRI \cite{Celiktutan:2019}  & HHI \& HRI & audio, physiological, and first-person vision & self-reports & \cmark  \\  \hline
\cite{Rudovic:2018,Park:2019} & HRI & \pbox{15cm}{facial expressions, body pose, audio\\(in children's storrytelling and therapy with robots)} & -- & \cmark \\ 

\bottomrule
\end{tabular}
\end{adjustbox}

\label{table:1}
\end{table*}

\section{Related Work}\label{section:related_work}
In recent years, the use of automated methods in classroom behavior analysis and engagement estimation has been on the rise. The popularity of such methods is largely due to the availability of big data and the progress of artificial intelligence. Notably, developments in deep learning have yielded significant results in social signal processing problems, including classroom and learning analytics.

We can categorize the literature of automated engagement estimation based on the following criteria:
\begin{itemize}
    \item learning situation (computer-enabled settings, classroom: traditional formation vs. group-work, etc.)
    \item nonverbal features (various behavioral cues can be related to learning-related activities.)
    \item computational methodology (in both feature extraction and machine learning)
    \item final objectives (showing a statistical relation vs. fully automated predictive system, psychologically valid measurements of engagement).
\end{itemize}

In addition to these points, another consideration is the use of sensors \cite{D'Mello:2017}. Whereas sensor-free measurements depend on intelligent tutor systems' log files, sensor-based measurements use physical devices such as physiological sensors (i.e., EDA, EEG, heart rate sensors) and audiovisual recordings acquired from cameras and voice recorders. As our motivation is to measure engagement as seamlessly as possible without necessitating any expensive and intrusive sensors, we limit our scope to engagement analysis using only visual modalities. Table~\ref{table:1} summarizes the literature of automated engagement analysis across three domains: classroom, computer-based settings (including intelligent tutors and screen-based learning games), and human-human, human-robot interactions (HHI/HRI). In the following subsections, we will review engagement studies in these three categories.

\subsection{Learning Analytics in the Classroom}
Despite the popularity of computer-based learning technologies, Intelligent Tutor Systems (ITS), and Massive Online Open Courses (MOOC), traditional classroom-based learning is still the dominant setting for primary through tertiary education. The popularity of classroom-based learning is primarily due to the importance of human factors and collaboration throughout the learning process. For this reason, analytics tools in the classroom that measure students' learning-related behaviors and affective and cognitive engagement may play an essential role for research aiming at improving the efficiency of classroom-based learning.

Learning analytics methods in the classroom may include video cameras in the corner of the room, direct recordings several students' faces and upper bodies, and external audio recorders. The quality of audio-visual feature extraction, in general, is not as fine-grained as in computer-based situations where a webcam, 1-2 meters away, captures a student's behaviors. However, classroom analytics provide more insight into student-teacher, student-learning material, and student-student interactions. 

To the best of our knowledge, Bidwell and Fuchs \cite{Bidwell:2011} proposed the first classroom monitoring system capable of analyzing student engagement. Although their technical report did not incorporate any quantitative results, they defined a general workflow of classroom analytics by using several color and Kinect depth-sensing cameras during a lesson in a third-grade classroom. Three observers attended the lesson and coded each students' behavior using a mobile device during 20 second intervals according to the following categories: appropriate (engaged, attentive, and transition) and inappropriate (non-productive, inappropriate, attention-seeking, resistant, and aggressive). Due to the limitations of only recording a single lesson and collecting highly imbalanced data, Bidwell and Fuchs used a Hidden Markov Model (HMM) to classify three categories (engaged, attentive, and transition) from head pose based gaze-target mappings.

A more recent classroom monitoring system was proposed by Raca and Dillenbourg \cite{Raca:2013:a}. Two ideas proposed in their study were to use students' motion information during class and student orchestration between neighboring students' feature representation to estimate student attention. In \cite{Raca:2015:dissertation}, they handcrafted several features such as eye contact (the percentage of time where faces are detected), amount of still time (where head pose does not change significantly for a period), and head travel (normalized head pose change). As ground truth labels of attention, Raca and Dillenbourg used self-reports that students completed in approximately 10-minute intervals. These features, together with a Support Vector Machines (SVM) classifier, performed up to the accuracy of $61.86\%$ (Cohen's $\kappa=0.30$) to predict 3-scale attention (low, medium, and high). Their seminal work showed that student attention can be automatically measured using visible behavioral cues. However, they used considerably long intervals (10 minutes) before self-reports were obtained. Moreover, they employed only attentional features (head pose and motion), not any affective or behavioral nonverbal features.

Zalatelj and Kosir \cite{Zaletelj:2017:a, Zaletelj:2017:b} used a Kinect sensor and its commercial SDK to estimate body pose, facial expressions, and gaze. Subsequently, they computed behavioral cues (i.e., yawning, taking notes, etc.) from Kinect features and trained a bagged decision tree classifier to estimate observer-rated attention levels (low, medium, high). However, some nonverbal features were extracted using Kinect's commercial SDK. They also used manually-labeled behavioral features (i.e., writing, yawning, one hand's touching head). Their experimental results included only a few minutes of video recording and the number of participants in their data sets was 3 and 6 students, respectively. Additionaly, the effect range of Kinect and similar depth sensors is around 1 to 3 meters. In a typical classroom with 20-30 students, several sensors are required, potentially introducing additional cost and device synchronization issues.

Thomas and Jayagopi \cite{Thomas:2017} collected video recordings of 10 students during three 12-minute intervals while they were listening to motivational video clips on YouTube. Three observers labeled the engagement of each student during 10-second intervals. They rated based on whether a student was looking towards the screen (teacher area), talking to a neighbor, or gazing in another direction. Their approach was to use head pose, gaze direction, and facial action unit features with SVM and logistic regression. The main limitation of this study was the limited data size and the lack of engagement labeling methodology. When a student is listening to the audio, looking to a voice source should not be considered a cue of attentiveness. Students can still focus on content when looking around or taking notes.

Goldberg et al. \cite{Goldberg:2019} is the first study that utilizes a psychologically valid and comprehensive engagement rating system. Their continuous observer-based rating system combines Chi \& Wylie's ICAP (Interactive, Constructive, Active, Passive) framework \cite{Chi:2014} and on-task/off-task behavior analysis \cite{Helmke:1988}. Using attentional (head pose and gaze direction) and affective (FACS action unit intensities) sets of features, as well as SVR, they predicted continuous observer-ratings and additionally showed the correlation between estimated engagement levels and self-reports collected at the end of 40-minute teaching units (N=52). They also showed that behavioral synchrony with immediate neighbors improved the estimating of engagement. 

One of the main objectives of learning analytics in the classroom is to report the estimated attention and engagement of students to teachers. For instance, Fujii et al. \cite{Fujii:2018} estimated head-down (i.e., taking notes or reading learning material) and head-up (gazing at whiteboard/teacher area) states for each student and depicted color-coded visualization to teachers with a sync rate. However, they tested the performance of the head-down/head-up detector on limited data. Also, reporting behavioral cues (looking at learning material or the teacher area) instead of engagement levels leaves teachers with minimal information. 

Two recent studies \cite{Ahuja:2019, Anh:2019} developed smart classroom monitoring systems. Whereas Anh et al. \cite{Anh:2019} used only gaze mapping and visualized the distribution on a dashboard, Ahuja et al. \cite{Ahuja:2019} integrated various nonverbal features in their smart classroom, EduSense. These features included the state-of-the-art methods in face detection and alignment, body pose estimation, hand raise detection, and active speaker detection.  \cite{Ahuja:2019} presented a technical analysis of real-time classroom monitoring systems, including the speed and latency of the system and algorithms' performance. However, they did not report on student engagement. Even though nonverbal features are essential to understand engagement, they are not easy to interpret, on their own, by a teacher.

In summary, computer vision-based classroom analytics studies are still limited. The sample sizes are small and the majority do not estimate attention or engagement levels. Besides, in studies that estimate student attention/engagement, there is no consensus regarding the assessment instrument.

\subsection{Engagement Estimation in Computer-based Learning}
Computer-based learning situations are more restricted than classroom situations because they only contain student-learning material interactions. These situations capture video and audio from 1 to 2 meters away, resulting in better quality feature extraction methods. Furthermore, introducing an intervention during learning is more straightforward than in the classroom setting. For these reasons, automated engagement estimation is more prevalent in computer-based situations where the participant plays an educational game, conducts reading comprehension or writing tasks, or learns with an ITS.

The first study we reviewed that predicts the level of engagement in computer-based settings (during which the participants perform a cognitive training task) was conducted by Whitehill et al. \cite{Whitehill:2014}. They used appearance-based facial features (Box filters, Gabor filters, CERT FACS features) and estimated levels of engagement using several classifiers such as GentleBoost, SVM, multinomial logistic regression. They developed a manual rating system (4-scales) and annotated the video recordings at 60-sec or 10-sec intervals. The accuracy of their classifiers vary between 36-60\%. 

In a similar computer-based setting, a writing task, Monkarasi et al. \cite{Monkaresi:2017} estimated engagement using Kinect face tracker ANimation Unit (ANU) features, LBP-TOP, and heart rate (estimated from videos of the face). They used concurrent self-reports (during the writing task every 2 minutes) and retrospective self-reports after the participants finished the task. Both self-reports showed high correlation ($r=0.82$, $p\le0.001$). Bosch et al. \cite{Bosch:2016:b} used estimated FACS action units as features and predicted BROMP annotations \cite{Ocumpaugh:2015} using different classifiers (Bayes Net, Updateable Naive Bayes, Logistic Regression, AdaBoost, Classification via Clustering, and LogitBoost). 


Another factor that plays an essential role in task engagement is mind wandering (MW). MW is defined as an attentional shift from the primary task and a subsequent decrease in task engagement \cite{Smallwood:2006}. For instance, in the learning context any thoughts that arise either intentionally or from boredom can be MW and linked to models of engagement.

The availability of automated methods to predict MW can reveal the covert aspect of engagement. The use of visual modalities, particularly face videos, to detect MW is preferable to eye gaze \cite{Smilek:2010} and physiological signals \cite{Blanchard:2014} which necessitate specialized sensors or textual features \cite{Mills:2015} that may depend on speech recognition, NLP, or manual labeling and relevant to our study. Swewart et al. \cite{Stewart:2017:a} is the first study that used the visual modality, facial action units and body motions to detect MW. They recorded facial videos while the participants watched a narrative film for 35 minutes. Each participant annotated MW by pressing a key through the screening. Facial action unit features and classifiers including logistic regression, naive Bayes, and support vector machines could spot MW in a person-independent setting with $F_1$ score of $.390$. Later, \cite{Stewart:2017:b} showed the generalizability of MW detection when trained and tested on different tasks (reading scientific text and watching a narrative film). Bosch et al. \cite{Bosch:2019} showed the applicability of MW detection in a classroom study (N=135) learning from an intelligent tutor system.
 
\subsection{Human-Human and Human-Robot Interactions (HHI/HRI)}
Another line of work is the attention analysis of human-human interactions, i.e., in group work and human-robot interactions. For example, Sanches-Cortes et al. \cite{Sanchez-Cortes:2011} proposed an audiovisual corpus of groups with four participants during a survival task and focused on estimating group performance, apparent personality, and perceived leadership and dominance.  Similarly, Rinvegal et al. \cite{Ringeval:2013} used a survival task during remote collaboration using audio, video, and physiological signals as well as self-reported engagement. However, although survival tasks can be useful to measure group interactions, they do not represent typical learning situations.  

Looking into more recent studies, Celiktutan et al. \cite{Celiktutan:2019} collected an audiovisual dataset during human-human and human-robot interactions using first-person cameras. They acquired self-/acquaintance-assessed personality and self-reported engagement labels. However, drawbacks that limit their setting include the scale of the dataset and interactions wherein one participant or robot asks predefined, standard questions. Another application in human-robot interactions is autism therapy for children \cite{Rudovic:2018,Rudovic:2019:a} and child-robot interactions (a dialogic storytelling task) \cite{Park:2019,Rudovic:2019:b}. The distribution of engagement during children's storytelling or autism therapy is more obvious and, in these settings, it is comparably easier to differentiate between engaged and disengaged behaviors than it is in schools where most pupils learn to hide their disengagement. Despite the lack of expert-labeling criteria, these studies adopt a continuous engagement labeling approach as in our engagement annotations. Furthermore, they used deep Q learning to actively sample training data and personalize models with limited data.  
 
To summarize, the literature in attention and engagement analysis is centered on computer-based learning settings as well as human-human and human-robot interactions. Collecting data for automated analysis in those domains is comparably more convenient than in the classroom. However, the impact of schools and classroom instruction exceeds the scope of these applications and, moreover, plays a crucial role in every student's life. Therefore, research on analyzing attention and engagement in the classroom is of high importance and may benefit from novel analytic approaches. Existing classroom-based studies are very limited in terms of data size. They were mostly conducted on university-level courses or on a small number of participants (mainly to test computer vision systems). While Raca and Dillenbourg \cite{Raca:2015:dissertation} conducted the most comprehensive attention monitoring study in the classroom and, thusly, showed the applicability of these technologies in a school setting, their study lacked expert-labeled attention/engagement measures and predictive learning models on a larger scale.
 

\section{Data Collection for Automated Engagement Estimation in the Classroom}\label{section:data_collection}
The study was conducted during regular lessons at a secondary school in Germany over a one and a half month period. The ethics committee from the Faculty of Economics and Social Sciences of the University of Tübingen approved our study procedures (Approval \#A2.5.4-097\_aa), and all teachers and parents provided written consent for their kids to be videotaped. Students who refused to be videotaped attended a parallel session covering the same instructional content. 

\subsection{Participants}
We collected audio-visual recordings of 47 classes from $5^{th}$ to $12^{th}$ grades, including 128 participants overall. Each participant attended more than one class ($3.84$ on average). Therefore, the total number of samples across grades was over 360. The collection of labelled data for developing and benchmarking automated methods is time-consuming. Thus, we identified a sub-sample of students based on their occurrence and visibility in multiple video recordings and used a sub-sample of 15 students from grade 8 (N=7) and grade 12 (N=8) in our analysis. Each participant appears five times on average and the total number of samples in our data is 75. Classes cover a wide range of subjects, including Mathematics, Chemistry, Physics, IMP (\textbf{I}nformatics, \textbf{M}athematics, \textbf{P}hysics), History, Latin, French, German, and English. 

\begin{figure}[h!]
\includegraphics[width=\columnwidth]{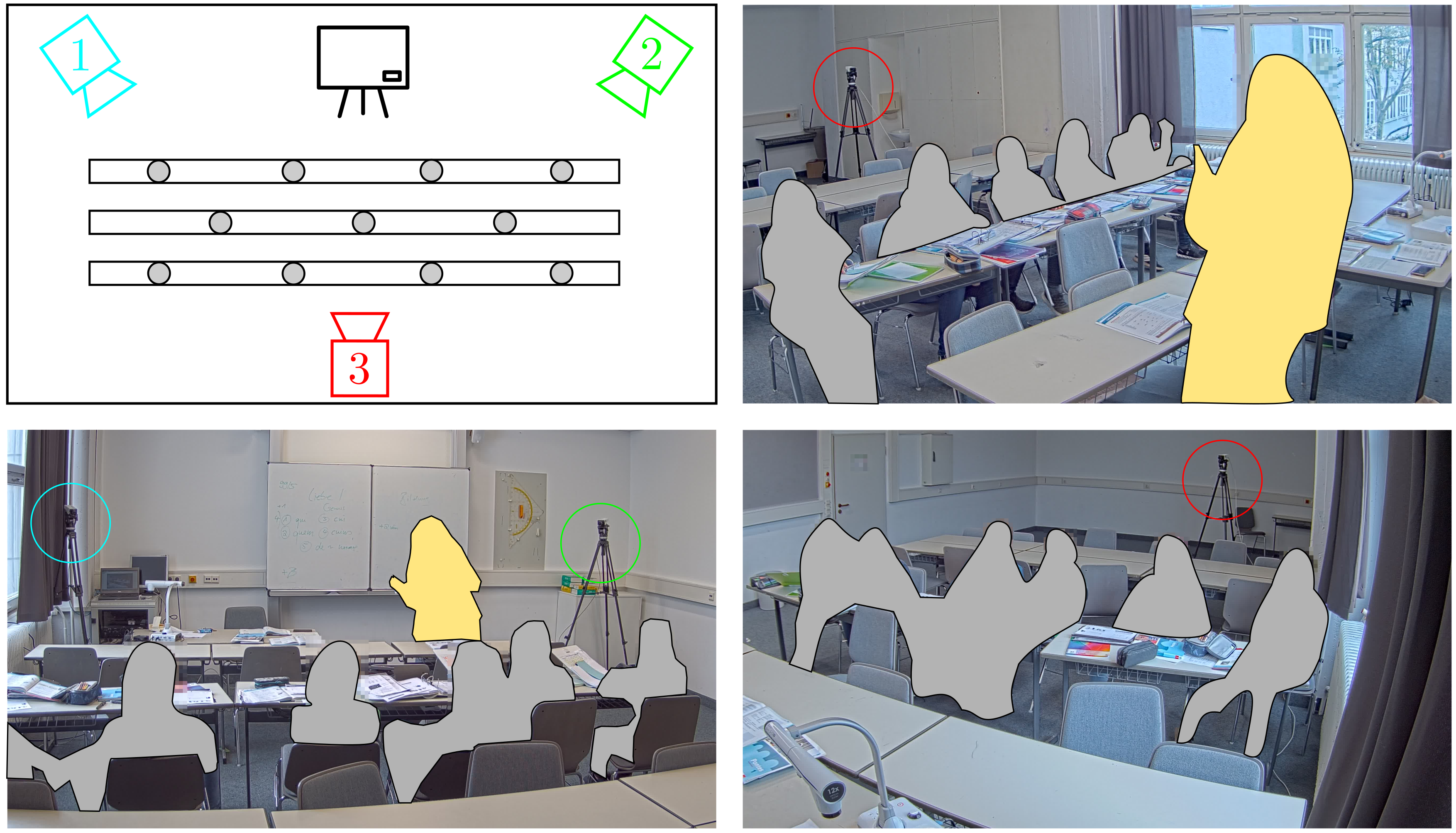}
\caption{Sample scene from the classroom. The synchronous cameras recorded the instruction simultaneously.}
\label{figure:sample_images}
\end{figure}

\subsection{Procedure}
Before classes on the first day, students filled out a questionnaire covering demographic information (age, gender) and individual prerequisites (BFI-2 XS, 15 items; \cite{Soto:2017}). After each session, students completed another questionnaire about their learning activities. Session recordings lasted between 30 and 90 minutes each. Video material during classes covered group work, individual work, and teacher-centered instruction. To best capture student attention on the instructor, we focused on teacher-centered components of the video (see Fig~\ref{figure:sample_images}), extracting the main part of instruction time in intervals of 15 to 20 minutes from each recording. The intervals were manually annotated by human raters.

\subsection{Self-Reported Learning Activities}
After each session, we assessed students' involvement (four items, $\alpha= 0.73$; \cite{Frank:2015}), cognitive engagement (six items, $\alpha = 0.78$; \cite{Rimm-Kaufman:2015}), and situational interest (six items, $\alpha = 0.92$; \cite{Knogler:2015}) during the preceding instructional period.

\subsection{Continuous Manual Annotation}
\begin{figure}[ht!]
    \centering
    \includegraphics[width=\columnwidth]{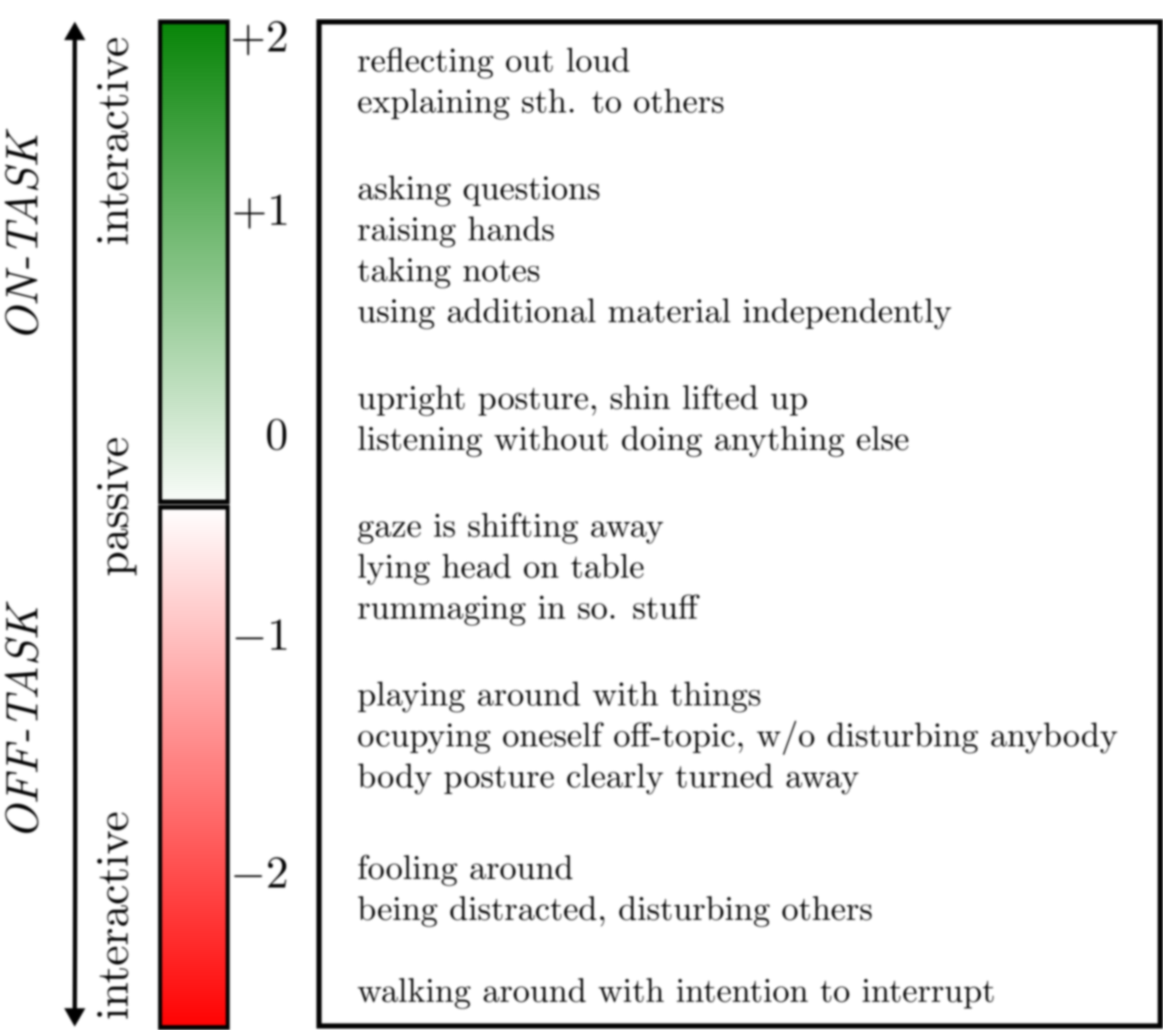}
    \caption{Continuous scale of our manual rating instrument and visible behavioral indicators \cite{Goldberg:2019}}
    \label{figure:manual_labeling}
\end{figure}

To manually annotate students' observable behavior, we used a one-dimensional scale in steps of seconds through the open software, CARMA \cite{Girard:2014}, which enables continuous interpersonal behavior annotation via joystick \cite{Lizdek:2012}. We also combined the concept of on-task/off-task behavior \cite{Helmke:1988, Hommel:2012} with existing scales from the engagement literature. To define more fine-grained cues within the possible behavioral spectrum, \textbf{I}nteractive, \textbf{C}onstructive, \textbf{A}ctive, and \textbf{P}assive, we gained inspiration from the ICAP framework \cite{Chi:2014}. Thus, behaviors were annotated on a symmetric scale ranging from -2, indicating disturbing (i.e., interactive), off-task behavior, to +2, indicating highly engaged, interactive, on-task behavior (see Fig~\ref{figure:manual_labeling}). Values closer to 0 indicated rather unobtrusive, passive behavior. Two raters annotated the sub-set of students in all videos in random order, with inter-rater reliability ICC(2,2) for each student being 0.77 on average (absolute agreement). For subsequent analysis, the mean across the two raters is calculated for every learner in every second. For more details about the manual annotation instrument, interested readers are referred to Goldberg et al. \cite{Goldberg:2019}. 

\subsection{Preprocessing}\label{subsection:preprocessing}
In each video recording, we had three cameras as depicted in Figure~\ref{figure:sample_images}. One camera was located in the rear part of the class covering the classroom and teacher and the other two cameras were placed on the left and right side of the teacher area (whiteboard) directed towards the class. We applied our computational pipeline to both the left and right camera and dynamically picked the stream where a particular student was more visible.

We used a single-stage face detector, RetinaFace \cite{Deng:2019}, to detect all faces in the video streams. Subsequently, we picked several query face images that belonged to the students whose behaviors we intended to analyze. Instead of face tracking, we directly used those query images and extracted ArcFace embedding \cite{Deng:2018} for all face patches. By calculating the minimum cosine similarity between the query images and all faces, we created face tracklets for each student. Despite the challenges of occlusion and different camera angles, the face detection and recognition methods we employed could localize and recognize faces most of the time due to their training on large and unconstrained data sets. We used one-second (24 frames) continuous sequences where both face detection and recognition worked smoothly.

Table~\ref{table:classroom_data_distribution} shows the number of different day recordings per student and the total length of the data where preprocessing worked. The total data length is 25,450 and 32,755 seconds in Grades 8 and 12. 
This amount makes over 15 hours of recording in 30 sessions. Compared to other classroom-based studies, the line of work by Raca \& Dillenbourg \cite{Raca:2015:dissertation} used four classes in 9 sessions. Even though their study was on large-scale data, their attention analysis was based on 10-minute intervals and self-reports. When we look into the size of other engagement studies in the classroom, the results are limited: three videos of 12-minute recordings in \cite{Thomas:2017}, 25 minutes of video recordings in \cite{Zaletelj:2017:a}, 4 minutes in \cite{Zaletelj:2017:b}.

\begin{figure}[ht!]
\includegraphics[width=\columnwidth]{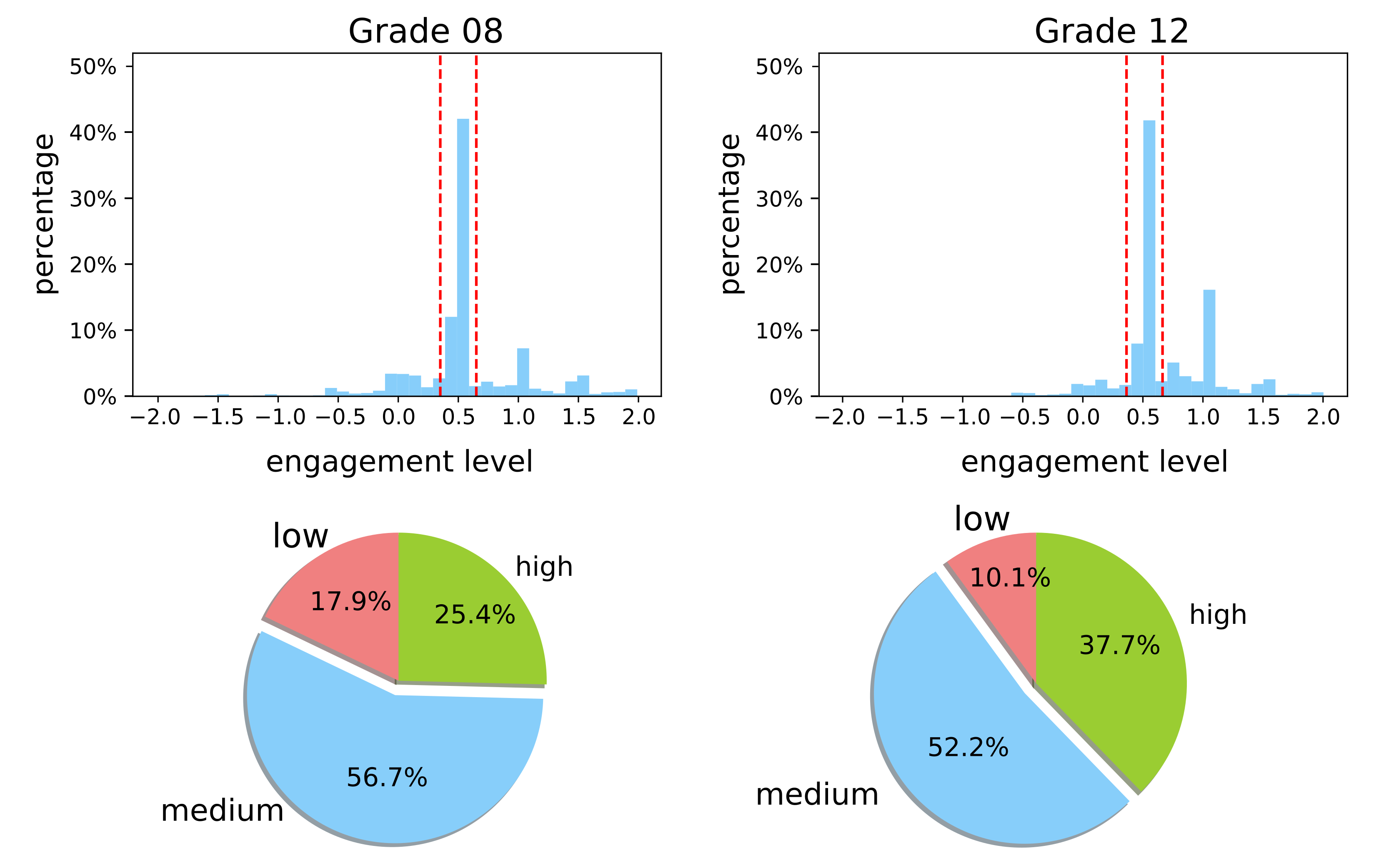}
\caption{The distribution of engagement labels in Grade 8 and 12. Pie charts show the percentage of quantized labels according to continuous labelling.}
\label{figure:histogram}
\end{figure}

In the continuous labeling scale, values denoting disengagement are rarely observed and the labels are often imbalanced. Thus, we followed the previous works that discretized the continuous scale into three scales: low [-2, 0.35], medium (0.35, 0.65], and high engagement (0.65, 2.0]. Figure~\ref{figure:histogram} depicts the continuous and discrete distribution of labels in Grades 8 and 12.

\section{Methodology}
\subsection{Problem Statement}
In our setting to classify engagement level, we used video recordings of classes. Formally, we employed sequences $\mathcal{S}=\{I_1,I_2,\cdots I_n\}$ where $n=1,\cdots,N$ denotes the time intervals of a second (24-frames). Using any of the modalities, we extracted feature vectors from each sequence $\mathbf{x}=\{x_1,x_2,\cdots x_n\}$ with $\mathbf{x} \in \mathcal{R}^{\mathcal{T}\times M \times D_m }$. The feature sequences are associated with engagement label $y=\{0,1,2\}$. To predict the engagement labels, we used either a single middle frame of a sequences or all frames in a temporal learning model.   

\begin{table}[ht!]
\caption{The number of classes and the total duration of recording where face detection works (in seconds) per each student.}
\begin{adjustbox}{max width=\linewidth}
\begin{tabular}{ l llllllll }
\toprule
\textbf{Grade 8} & & & & & & & &   \\ \toprule
\rowcolor{Gray}
\textbf{student}  & \textbf{S4}  & \textbf{S7}  & \textbf{S8} & \textbf{S11}  & \textbf{S13}  & \textbf{S14}  & \textbf{S16} &  \\
\#class & 2  & 7  & 7 & 3  & 6  & 4  & 6 &   \\
\rowcolor{Gray}
seconds & 836  & 5450  & 5309 & 2269  & 4404  & 2674  & 4508 &  \\ \toprule
\textbf{Grade 12} & & & & & & & &   \\ \toprule
\rowcolor{Gray}
student  & S1  & S2  & S3 & S4  & S5  & S6  & S7 & S8  \\ 
\#class & 9  & 8  & 3 & 3  & 4  & 3  & 6 & 4   \\
\rowcolor{Gray}
seconds & 6363  & 6695  & 2662 & 2708  & 4219  & 2605  & 3844 & 3659  \\ \bottomrule
\end{tabular}
\end{adjustbox}
\label{table:classroom_data_distribution}
\end{table}

\subsection{Feature Representation}
In most of the classes, students were listening to the teacher instead of speaking. Due to occlusion of the students' upper bodies in many of the recordings, nonverbal features such as speech and body pose are not always available. However, faces are usually visible and computationally faster and more reliable to detect. Consequently, our analysis depends on preprocessed faces as described in \ref{subsection:preprocessing}.

Motivated by the fact that engagement is a multidimensional construct, we can extract two different sets of information from face images: attentional and emotional features. There are several studies in the literature that used available face processing tools such as OpenFace \cite{Baltrusaitis:2018} for engagement estimation \cite{Rudovic:2018,Kaur:2018}. 
\begin{figure}[b!]
    \centering
    \includegraphics[width=\columnwidth]{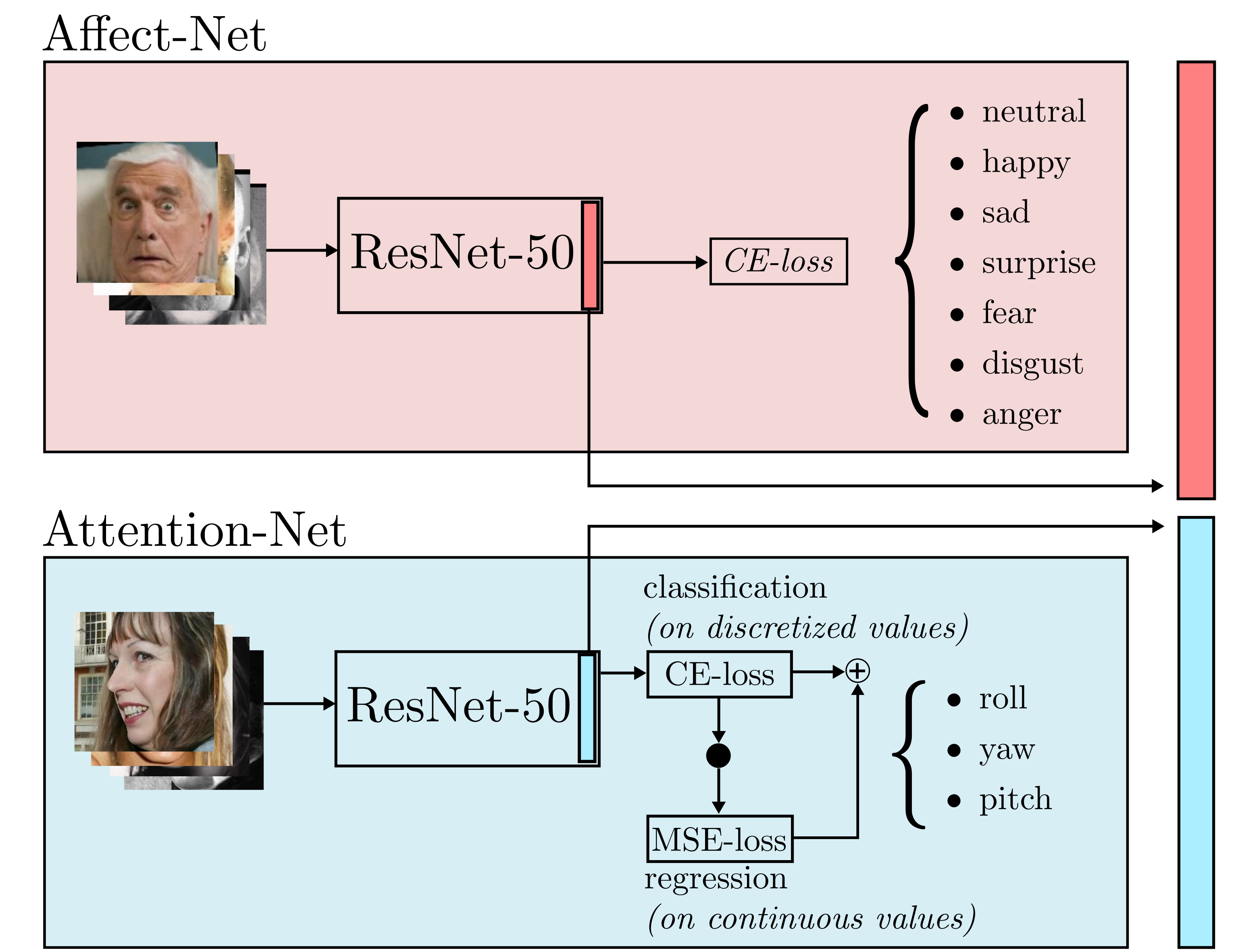}
    \caption{Feature learning for affect and attention. Two ResNet-50 backbones are separately trained for facial expression recognition and head pose estimation. The learned features subsequently will be used for engagement estimation on classroom data.}
    \label{figure:feature_extraction}
\end{figure}

The main drawback of this approach, however, is that it depends on very accurate face alignment. In the classroom, camera distance from students varies between 2-10 meters, and this reduces image quality and eventually leads to poor facial keypoint localization. When we processed the classroom data using \cite{Baltrusaitis:2018}, it could process approximately 30-40\% of a students' face in a class with high confidence. Furthermore, even though facial action unit-based approaches provide valuable information on affect, they almost always anticipate nearly frontal images. Considering these issues, we extracted affect features based on categorical facial expression recognition and attention features based on head pose estimation without depending on 68-point facial landmarks.

Figure~\ref{figure:feature_extraction} shows the feature learning for affect and attention. In affect branch (Affect-Net), we used one of the most unconstrained and large-scale affect datasets, AffectNet \cite{Mollahosseini:2019}, and trained a ResNet-50 network using softmax cross entropy loss to predict categorical models of affect (seven discrete facial expressions): neutral, happy, sad, surprise, fear, disgust, and anger. The training set of AffectNet is composed of 23,901 images, whereas the validation set has 3,500 images. We aligned all face images using five facial keypoints that were estimated by face detector \cite{Deng:2019} and aligned by similarity transform to the size of $224\times224$. The training was done using an SGD solver with an initial learning rate of 0.1 (decayed ten times in each 30 epochs) for 100 epochs. The best accuracy on the validation set reached 58\%. We used the layer's feature activations before the last fully connected layer as affect embedding.

We used another ResNet-50 backbone (Attention-Net) to learn attention features. By adopting the approach in \cite{Ruiz:2018}, we trained the network on 300W-LP \cite{Zhu:2016} to estimate head pose jointly by softmax classification on discretized values and mean squared loss on continuous values. The advantage of the CNN-based approach for head pose estimation is that it is more robust than Perspective n-Point (PnP)-based methods that find correspondence between estimated facial keypoints on image and their corresponding 3D locations in an anthropological face model. In challenging cases where those methods fail, CNN-based methods can return satisfactory predictions and, more importantly, map the inputs in a continuous low-dimensional embedding according to poses.

As the training corpus is very large and contains various challenging situations in both Affect-Net and Attention-Net branches, these methods learn robust features. Compared to the handcrafted appearance features such as Local Binary Patterns or Gabor filters, deep embeddings can be extracted without precise alignment and are extendable by training with new DNN architectures on more data. We trained Attention-Net and Affect-Net representations on head pose estimation (300W-LP) and facial expression (AffectNet) datasets. To avoid overfitting due to the limited number of subjects represented in the classroom data, we did not perform any finetuning on student engagement data. 

\subsection{Engagement Classification}
For both modalities, attention and affect, we trained several classifiers. In frame-based classifiers, we trained using only the middle frame of each 1-second sequence to avoid redundant training samples when all frames were used. We additionally reasoned that kernel-based methods take a longer time to train. In the test phase, we retrieved all 1-second (24-frame) sequences' predictions and applied majority voting. 

We built our models using classifiers in two categories: shallow classifiers, and Deep Neural Networks (DNN). Shallow classifiers that we used are Support Vector Machine (SVM), and Random Forests (RF) classifiers. All model training and dimensionality reduction were conducted in a person-independent manner. Considering the behavioral differences between grades, we did all experiments separately in Grade 8 and 12. 

In SVM, we tested linear SVM and also SVM with radial basis function (rbf) kernel. Training SVM-based models with a large number of instances and features (i.e., 2048-dimensional features and 20-25K training samples) increases required memory and training time. Thus, before training SVM models, we applied Principal Component Analysis (PCA) and used the principal components that explain 99\% of the variance in the corresponding training set. In this way, 2048-dimensional feature embeddings were reduced to the dimension of 48. In RF, we used feature embeddings directly without dimension reduction.  

In DNN's, the first approach is to use a Multi-Layer Perceptron (MLP). Here we discuss the variation of the data to emphasize why we only trained an MLP instead of retraining the entire representation up to the first layers of ResNet-50 architecture. Even though the data subset that we acquired for manual annotation and used in our analysis is over 15 hours, we still faced a problem due to the slow nature of classroom activities and limited number of subjects. Propagating the entire network results in an easy overfitting of the data and the failure to recall previously learned features useful for understanding engagement.

We used a two-layer MLP with a hidden layer in a size of 128. Training is done in mini-batches of 256 using soft-max cross-entropy loss and SGD solver with a learning rate of 0.001. In each trial, we kept a random 10\% of the training data as a validation set for early stopping. 
In both SVM and MLP models, we applied majority voting to acquire the prediction of 1-second sequences. In addition to those approaches, we used a recurrent neural network model, long short-term memory (LSTM) \cite{Sepp:1997}, to directly learn on temporal data. LSTMs showed great performance in modelling long-term dependencies in various problems such as language modelling \cite{Graves:2012}, neural machine translation \cite{Sutskever:2014}, visual recognition and video action recognition \cite{Donahue:2017}.

In contrast to feedforward neural networks, recurrent neural networks can learn from temporal data. In learning a problem, the memory cell of a recurrent network (here, we use LSTM cell) is defined not only by the current inputs but also by longer temporal dependencies. The key contribution of LSTMs is self-loops that produce paths through which gradients can flow, reducing the chance of an exploding or vanishing gradient. LSTMs are controlled by a hidden unit and the integration of the time scale can change. 


We provided 2048-length Attention-Net or Affect-Net embeddings as input to a two-layer LSTM network with a hidden size of 128. The output of the LSTM network on the last time step is fed to a fully connected layer in size of 64, and the entire model is trained using softmax cross-entropy loss and Adam solver \cite{Kingma:2015} with a learning rate of 0.001. All LSTM models are trained for 5 epochs.

\begin{figure}[ht!]
    \centering
    \includegraphics[width=\columnwidth]{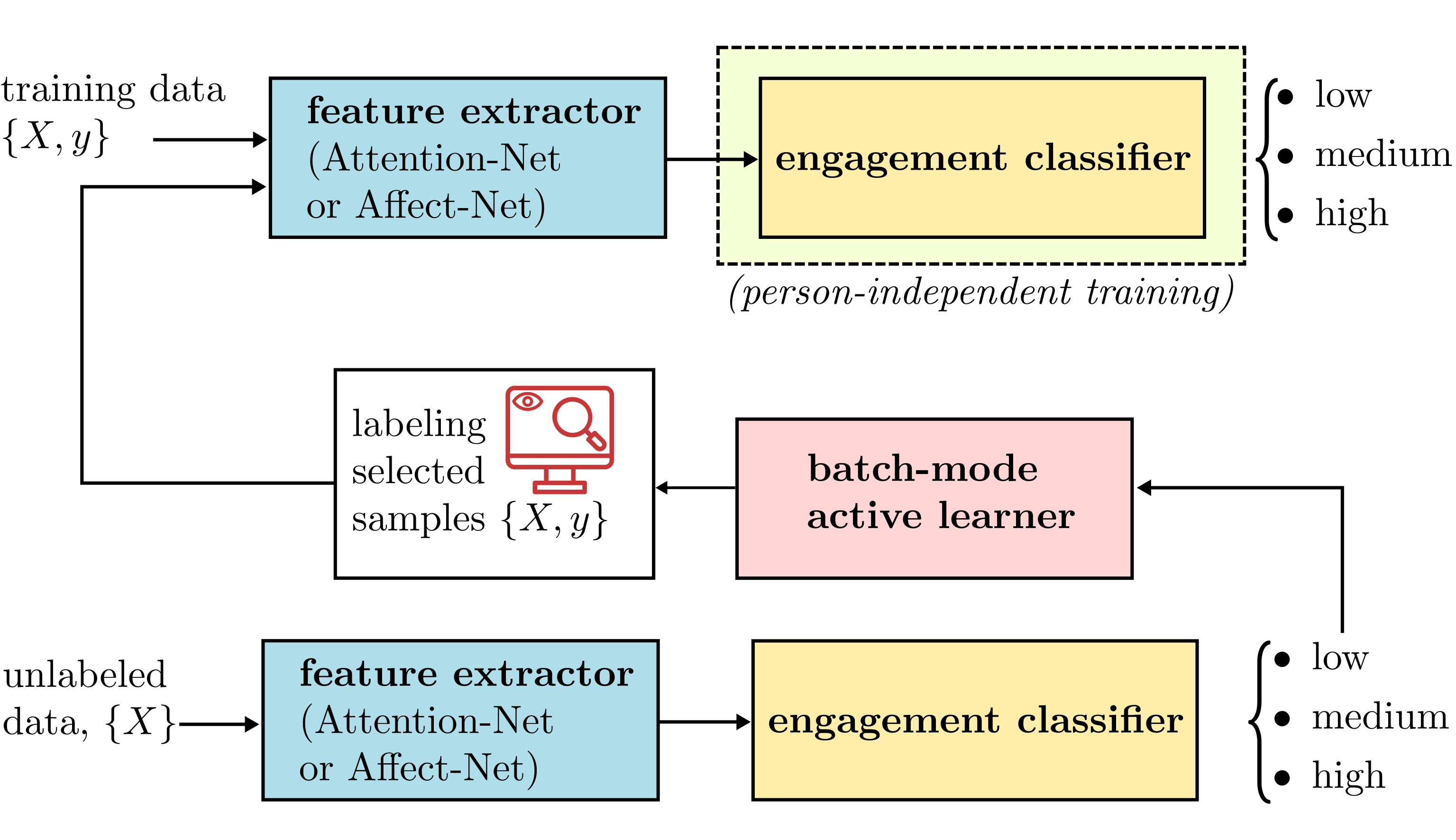}
    \caption{Batch-mode Active Learning for Personalized Engagement Classification (The initial network is the engagement classifier trained in a person-independent manner and the weights of the feature extractors kept frozen during all experiments).}
    \label{figure:active_learning}
\end{figure}

\begin{table*}[ht!]
\caption{Performance Comparison of Engagement Classifiers on Classroom Data using Attention-Net and Affect-Net Features and Different Classifiers.}
\label{table:resultsA}
    \centering
    \begin{tabular}{lll l ll}
        \toprule 
        \textbf{Classifier} & \multicolumn{5}{c}{\textbf{AUROC}}  \\
        \toprule 
                            & \multicolumn{2}{c}{Grade-8}  & &  \multicolumn{2}{c}{Grade-12} \\
                             \cmidrule{2-3} \cmidrule{5-6}
                            & \emph{Attention-Net} & \emph{Affect-Net} & & \emph{Attention-Net} & \emph{Affect-Net}\\
        \midrule
        SVM (linear) &  .560 $\pm$ .05 & .570 $\pm$ .06  & & .656 $\pm$ .09 &   .563  $\pm$ .06  \\
        SVM (rbf)    & .603 $\pm$ .05  & .604 $\pm$ .03 & &  .697 $\pm$ .07  & .595 $\pm$ .08  \\
        RF           & .620 $\pm$ .04  & .608 $\pm$ .03 & & .708 $\pm$ .05 & .600 $\pm$ .09  \\
        MLP          & .615 $\pm$ .05  & .597 $\pm$ .03  & & .701 $\pm$ .06 & .622 $\pm$ .05 \\
        LSTM         & .603 $\pm$ .05  &  .610 $\pm$ .04 & & .719 $\pm$ .05 & .612 $\pm$ .09  \\
        \bottomrule
    \end{tabular}
\end{table*}

\subsection{Personalization of Engagement Classifiers}
There are two plausible use cases for an engagement estimation system that can classify the engagement levels of all students in the classroom. Such a system could be used in classroom management studies or as part of an affective and cognitive interface to help teachers understand classroom engagement and regulate teaching styles accordingly. In order to be effective, the system needs to be used many times in the same classroom. Furthermore, engagement and disengagement during instruction can differ significantly from one student to another. Thus, engagement classifiers could benefit from personalization. 

The initial step is to train the engagement classifier on person-independent training data. In SVM-based classifiers, probability outputs can be calculated by cross-validation and Platt scaling, whereas the mean predicted class probabilities of the trees can be used in Random Forests. On the other hand, MLP and LSTM classifiers provide probability output because they were trained with softmax cross-entropy loss. These probabilities will be used to associate an uncertainty score to unlabeled instances.

Typically, traditional active learning algorithms propose a single instance to label at a time and this may result in a longer waiting time for the expert labeler during the personalization of the engagement classifier. We assume the labeler starts from an engagement classifier trained in a person-independent manner and labels a set of instances. In order to investigate the effect of personalization with a small amount of data, we utilize the margin uncertainty principle that considers the samples with the smallest margin between the first and second most likely class probabilities. These samples can be considered more difficult, and labeling them helps define a better separation among the engagement intensities. The margin uncertainty rule can be written as follows:
\begin{equation}
x_{marg}^* = \argmin_x \big[ P_{M_{init}}(\hat{y_1} \mid x) - P_{M_{init}}(\hat{y_2} \mid x) \big]
\end{equation} 
where $\hat{y}=P_{M_{init}}(\hat{y} \mid x)$ is the prediction with highest posterior probability, $\hat{y_1}$ and $\hat{y_2}$ are first and second most likely predictions.

Figure~\ref{figure:active_learning} depicts our personalization framework using batch-mode active learning. As there is no additional training in the deep embedding part (Attention-Net and Affect-Net) on engagement at that stage, training time is not increased. Only a small batch of unlabeled data is sent to the oracle in each episode, and the classifier part is retrained. Instead of updates with a single instance, we sampled a small batch of unlabeled images to label, removed them from the pool, and retrained the initial model iteratively. In this way, each personalization step is applied on a day or a week of recording, and a batch is composed of the most qualitative samples to adapt the existing engagement classifier on a specific subject.

\subsection{Results}\label{section:results}
As we report on the results of our work, it is important to note that we performed engagement classification experiments separately in grades 8 and 12 because visual engagement across grades may vary. In each grade, training and testing were conducted in a person-independent manner. With the exception of the test subject, every student in every grade was used in training and the same experiment was repeated per student, modality (affect vs. attention), and grade. Table~\ref{table:resultsA} shows the performance of various classifiers using Attention-Net and Affect-Net features. We used weighted Area Under the ROC Curve as a performance measure in the task 3 level engagement classification because it measures the performance of a classifier in different thresholds. Furthermore, it is more attune to class imbalances than metrics such as accuracy.

\begin{table}[b]
\caption{Performance Comparison of Different Fusion Strategies using Random Forest Classifiers.}
\centering
\begin{tabular}{ llc }
\toprule
\textbf{Grade} & \textbf{Feature Set} & Avg. AUROC \\
\midrule
8 & Attention-Net & .620 \\
8 & Affect-Net    & .608 \\
8 & Feature-level Fusion & \textbf{.633} \\
8 & Score-level Fusion & .632 \\
\midrule
12 & Attention-Net & \textbf{.708} \\
12 & Affect-Net    & .600 \\
12 & Feature-level Fusion & .616 \\
12 & Score-level Fusion & .694 \\
\bottomrule
\end{tabular}
\label{table:resultsB}
\end{table}

\textbf{Engagement classification.} The criteria for the manual annotation of engagement (as depicted in Figure~\ref{figure:manual_labeling}) is on a higher level and not directly related to gaze direction or facial expressions. When visual indicators were compared, Attention-Net features yielded .01 to .03 better AUC than Affect-Net in Grade 8. On the other hand, the margin between the average AUCs of Grade 12 students is more considerable; attention-net features performed .08-.11 better than Affect-Net features in Grade 12. This situation may be related to the easy distraction, movement, and increased gaze drifts characteristic of students in both grades. As a result, attention features capture engagement effective than affect features.

Another comparison is the type of classifier used to examine engagement. In the literature, shallow classifiers perform better than DNN methods in engagement and similar affective computing problems. In our experiments, linear SVM classifiers fall behind all other classifiers (.03 to .06 in AUC). However, there is no explicit performance gain among SVM with rbf kernel, RF, and MLP classifiers across both grades and feature sets. We would expect deep learning-based methods, for instance, MLP could capture engagement better than shallow classifiers, but it is comparable to RF and SVM-rbf. This may be due to the limited sample size of the data, the multifaceted aspect of learning problems, and imbalances in feature and label distribution. As we transfer feature representations of engagement from similar tasks and large-scale corpus, better feature representations facilitate engagement classification, and the margin among the classifiers is not wide. In overall performance, the best performing engagement classifiers are the ones that depend on attention features in Grade 12 (i.e., AUC of .708 and .719 with RF and LSTM classifiers).

Looking into DNN-based classifiers, the use of temporal information during training improved the performance of MLP only in the settings of Affect-Net/Grade 8 (+.013 in AUC) and Attention-Net/Grade 12 (+.018 in AUC). The limited improvement of LSTMs can be due to the short time window (24-frame). As our continuous engagement labeling approach gathers engagement labels per second and we aim to predict engagement per second, we stuck to the same setting in all experiments and predicted 1-second intervals at a time so as not to introduce delay and produce real-time feedback for the teacher when deployed in a school setting. 

\begin{figure*}[h!]
    \centering
    \input{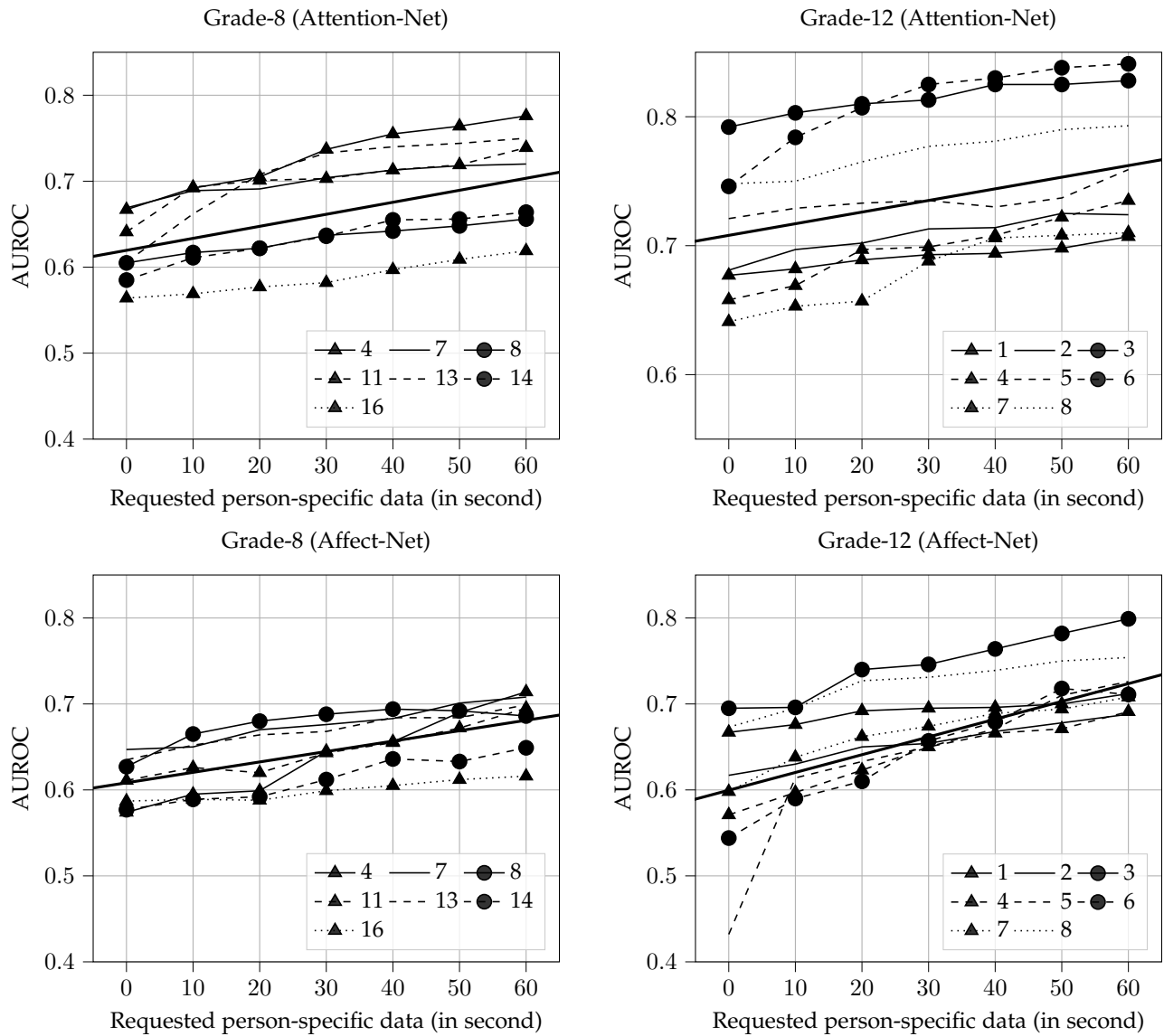}
    \caption{The Effect of Personalization on Different Engagement Classifiers (All classifiers are based on RF. The legends show the corresponding AUC performance per student, and each thick line represents the overall trend of personalization.)}
    \label{figure:results:personalization}
\end{figure*}

Besides the average AUC performance of different feature sets and grades, there are intrapersonal variations in the performance of engagement classification. A classifier performs better in a student using affect features, whereas attention features on another student using the same classifier outperform affect features. 

We tested different fusion strategies using RF engagement classifiers. Table~\ref{table:resultsB} shows the performance of feature-level and score-level fusion in Grade 8 and 12. In Grade 8, both fusion strategies yielded comparable improvement, +.012-.013 of AUC over the best performing modality (Attention-Net). On the other hand, score level fusion in Grade 12 is on par with the performance of attention features, whereas feature-level fusion performed slightly above Affect-Net performance.

\begin{table}[hb!]
\caption{Confusion Matrices for the Best Person-Independent and Personalized Models.}
\centering
\begin{tabular}{ lll lll }
\toprule
\textbf{Method} & \textbf{Actual} & \multicolumn{3}{l}{\textbf{Classified}} & \textbf{Priors}\\
\toprule
\emph{(Grade 12)}    &            & \emph{low} & \emph{medium} & \emph{high} &  \\
Attention-Net, RF  & \emph{low}    & .099 & .442 & .458 & .101 \\
                & \emph{medium} & .053 & .735 & .345 & .522 \\
                & \emph{high}   & .075 & .400 & .525 & .377 \\
\midrule
Attention-Net, RF  & \emph{low}  & .185 & .387 & .429 & .101 \\
(personalized)  & \emph{high}    & .027 & .768 & .205 & .522 \\ 
                & \emph{high}    & .032 & .360 & .608 & .377 \\
\bottomrule
\end{tabular}
\label{table:confmat}
\end{table}
\vspace{0.1cm}
Reviewing the overall results, the average AUC ranged between .560 to .719. The performance gap between Attention-Net and Affect-Net features was rather limited (.01-0.2) in Grade 8; however, Attention-Net features outperformed Affect-Net features by a large margin, +.08-.11 of AUC. When the difficulty of interpreting a student's intensity of engagement using only facial videos is considered, these results are satisfying. Our analysis validated that student engagement could be estimated independently from the grade and course content over a long period of time.

\textbf{Personalized models.} In the engagement classifier's personalization, we picked RF classifiers because of their successful performance in person-independent experiments and speed in training. Both SVM classifiers with rbf kernels and DNN models take a longer time to train.

\begin{figure}[ht!]
    \centering
    \includegraphics[width=\columnwidth]{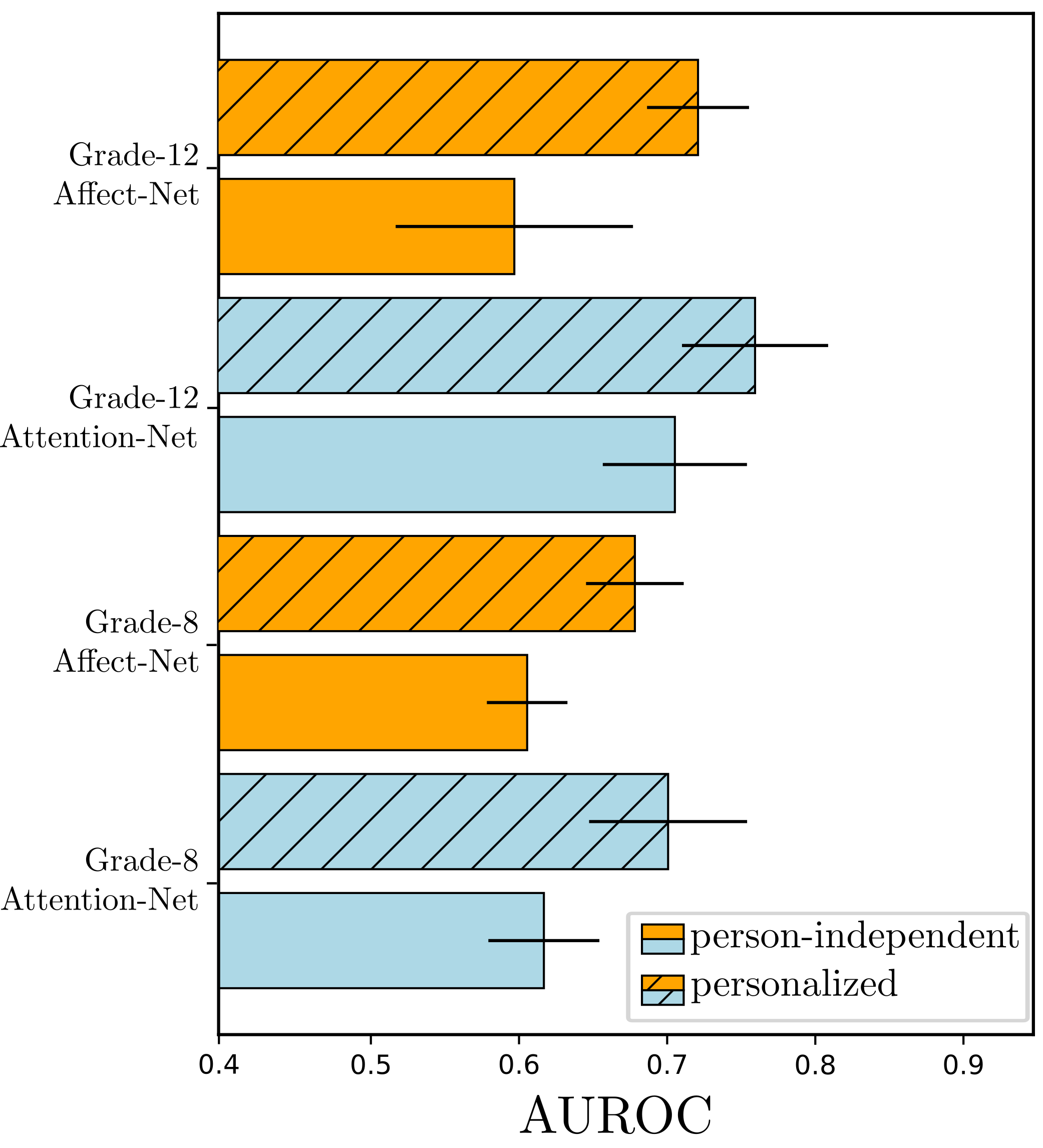}
    \caption{The overall improvement of personalization in AUROC using Attention-Net and Affect-Net features in Grades 8 and 12.}
    \label{figure:results:personalization2}
\end{figure}

Instead of directly training and testing on person-specific data, we adapted person-independent models and checked the effect of small person-specific data in an active learning setting. The number of samples from each student varied (as depicted in Table~\ref{table:classroom_data_distribution}). Thus, we limited person-specific data to be requested labels by the oracle as 60 seconds for each student.

The effect of personalization with RF engagement classifiers using Attention-Net and Affect-Net features in Grades 8 and 12 is depicted in Figure~\ref{figure:results:personalization} for each student individually. Similar to previous results, we reported Area Under the ROC Curve. In each experiment, we started from the model trained in a person-independent manner, sampled 60 samples using different sampling strategies, and compared ROC performance to the initial one. The 60 samples were acquired after 6 steps by selecting only 10 samples at a time, adapting the classifier with new samples, and continuing to use the samples iteratively. As the amount of data per student changes, 60 samples correspond to different ratios of the entire person-specific data; thus, we also reported the requested (\%) ratio of these samples to each student's total amount of data.

Except for one student (S4 in Grade 8), the amount of data is large enough, and the requested data (60 samples) corresponds only 2-3\% of the entire data. The effect of personalization varies from .03 to .29  of AUC. Affect features performed in both grades show greater improvement after personalization. The same amount of person-specific samples causes +6.89 and +9.83 of AUROC in attention and affect features, respectively.

Even though we limited the additional data introduced during personalization to 60-seconds, personalization helped up to +.12 of AUC. Table~\ref{table:confmat} shows the confusion matrices of RF classifier using Attention-Net features in Grade 12 before and after personalization. High engagement is misclassified mostly as medium (.400 and .360); however, low engagement's misclassification is medium and high. This can be due to the class imbalances. Of the few samples from low (10.1\% in Grade 12), the worse performing class is low engagement. On the other hand, the improvement in medium and high is clearer after personalization, .735 to .768 and .525 to .608.

Powerful feature representations, Attention-Net, and Affect-Net for face images facilitated further engagement classification. Our experiments in personalization showed that performance can be improved on average +.084 by using 60 seconds of personal data. The largest improvement, as depicted in Figure~\ref{figure:results:personalization2}, is +.124 of AUC in Affect-Net features and RF classifier in Grade 12.  

Labeling 60 one-second samples picked from different parts of a video is more manageable than labeling the entire recording and takes only a few minutes for an expert annotator. In return for this effort, the performance gain was substantial in both feature sets and grades. Thus, the personalization of engagement classifiers should be considered in large-scale classroom studies.


\section{Discussion}\label{section:discussion}

\subsection{Main Findings}

In contrast to the previous works that used mainly handcrafted local (i.e., local binary patterns, Gabor filters) and precomputed features such as head pose or estimated facial action units, we showed that engagement as a 3-class classification problem can be predicted in the classroom.

We gathered a large-scale classroom observation dataset and collected the observer ratings of student engagement for Grades 8 and 12 (N=15). In contrast to the limited training and testing protocols in the literature, our study is the first to validate the use of automated engagement analysis in the classroom. 

Our work proves that even a small amount of person-specific data could considerably enhance the performance of engagement classifiers. In comparison to the person-independent settings of many machine learning and computer vision tasks, personalization in engagement analysis significantly impacts performance. We find this to be the case because of personal differences in visible behaviors during levels of low and high engagement. Furthermore, engagement can even reveal variation in time (for instance, the indicators of engagement are not the same in different classes, i.e., math and history).

\subsection{Limitations and Future Work}
From the technical perspective, the limited sample size is related to both the technical constraints with regard to the infrastructure in such field studies (e.g. the preparation of such a recording requires 20 minutes) and the manual effort associated with preparation and post-processing. Additionally, the presence of cameras can put pressure on students and cause their behavior to change when they know instruction is being recorded. Collecting a significant scale of audiovisual recordings from the same classes over the course of a school year as longitudinal study could overcome these effects and allow researchers to investigate engagement in time.

Another limitation of this study is its focus on only the visible dimension of engagement. The detection of mind wandering through observation of a subject's face is a relevant emerging research topic. Combining automated methods to detect mind wandering with engagement analysis may offer a solution and yield a better understanding of students' affective and cognitive behaviors in the classroom.

Even though our study presents a step towards learning facial representations in the classroom, it was not possible to learn them on the engagement data due to the limited sample size. The use of self-supervision and representation learning on unlabelled classroom data may result in better representations for engagement analysis in future work. Additionally, our models failed to detect low engagement. The distribution of continuous labeling was also highly imbalanced. To solve these issues, we propose collecting more data in uncontrolled environments or, in order to obtain additional low engagement samples, employing interventions to manipulate engagement.

\subsection{Applications and Ethical Considerations}
The reported results in this study suggest that engagement classifiers could be applied to automate data processing within the scope of classroom instruction research and also personalized using a small amount of data. We are hopeful that automated engagement analysis becomes a research tool of classroom instruction research in the near future. In particularly, as part of responsible use of the data, our approach envisions beyond anonymization~\cite{sumer2020automated} the immediate deletion of the raw video recordings. Instead, only aggregated information of the student group might be stored, which could be used for investigation instructional quality. As a result, an explicit mapping to a student's individual engagement scores and features can be avoided. Further improvement in the performance of engagement classification and a transition from student engagement to classroom-level analysis has the potential to make engagement analysis a more superior tool.

Data collection, storage, and privacy concerns are the most significant issues that need to be addressed before large-scale classroom studies can be conducted. Labeling a small amount of personalized data improves the performance of engagement classifiers. Personalized models using the same feature extractor learned from the data can easily be applied to many real-time students. Instead of recording videos, such a system can record only behavioral data and substantially help increase the sample size for studies in classroom research.
 
We are well aware that a potential application for engagement classifiers is a real-time classroom observation systems such as \cite{Ahuja:2019}. Although such affective and cognitive interfaces summarizing engagement analytics as a teaching aid could help to enhance teaching quality, we strongly oppose any use of such solutions for real-world classroom monitoring for both ethical reasons and the lack of empirical data on possible negative side effects with regard to students' motivation and learning in such arrangements. Importantly, advances in machine learning should address the fairness, accountability, transparency, and bias of algorithms before being deployed in any application. In this context, only a continuous, reflective dialog with social stakeholders can lead to sustainable solutions.

\section*{Acknowledgments}
Ömer Sümer is a doctoral student at the LEAD Graduate School \& Research Network, which is funded by the Ministry of Science, Research and the Arts of the state of Baden Württemberg within the framework of the sustainability funding for the projects of the Excellence Initiative II. This work is also supported by Leibniz-WissenschaftsCampus
Tübingen ``Cognitive Interfaces''.

\bibliographystyle{IEEEtran}
\bibliography{IEEEabrv,main}
    
\begin{IEEEbiography}[{\includegraphics[width=1in,height=1.25in,clip,keepaspectratio]{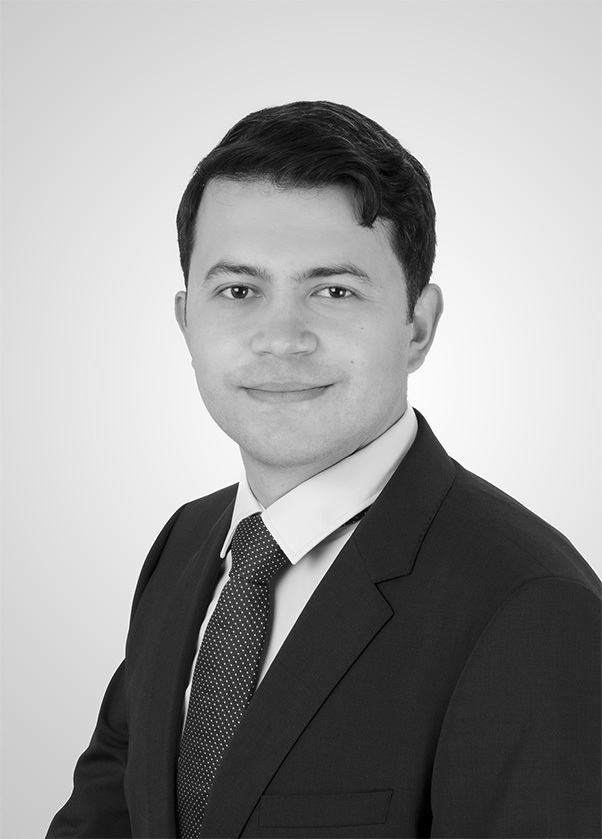}}]{Ömer Sümer}
received the BSc degree in electronics engineering from Naval Academy in Istanbul, Turkey, and the MSc degree in electronics engineering from Istanbul Technical University in Istanbul, Turkey. As of August 2017 he started the PhD
degree in computer science in the Group of Human-Computer Interaction at the University of Tübingen. His research interests involve machine learning, computer vision and their application in social and affective computing. 
\end{IEEEbiography}

\begin{IEEEbiography}[{\includegraphics[width=1in,height=1.25in,clip,keepaspectratio]{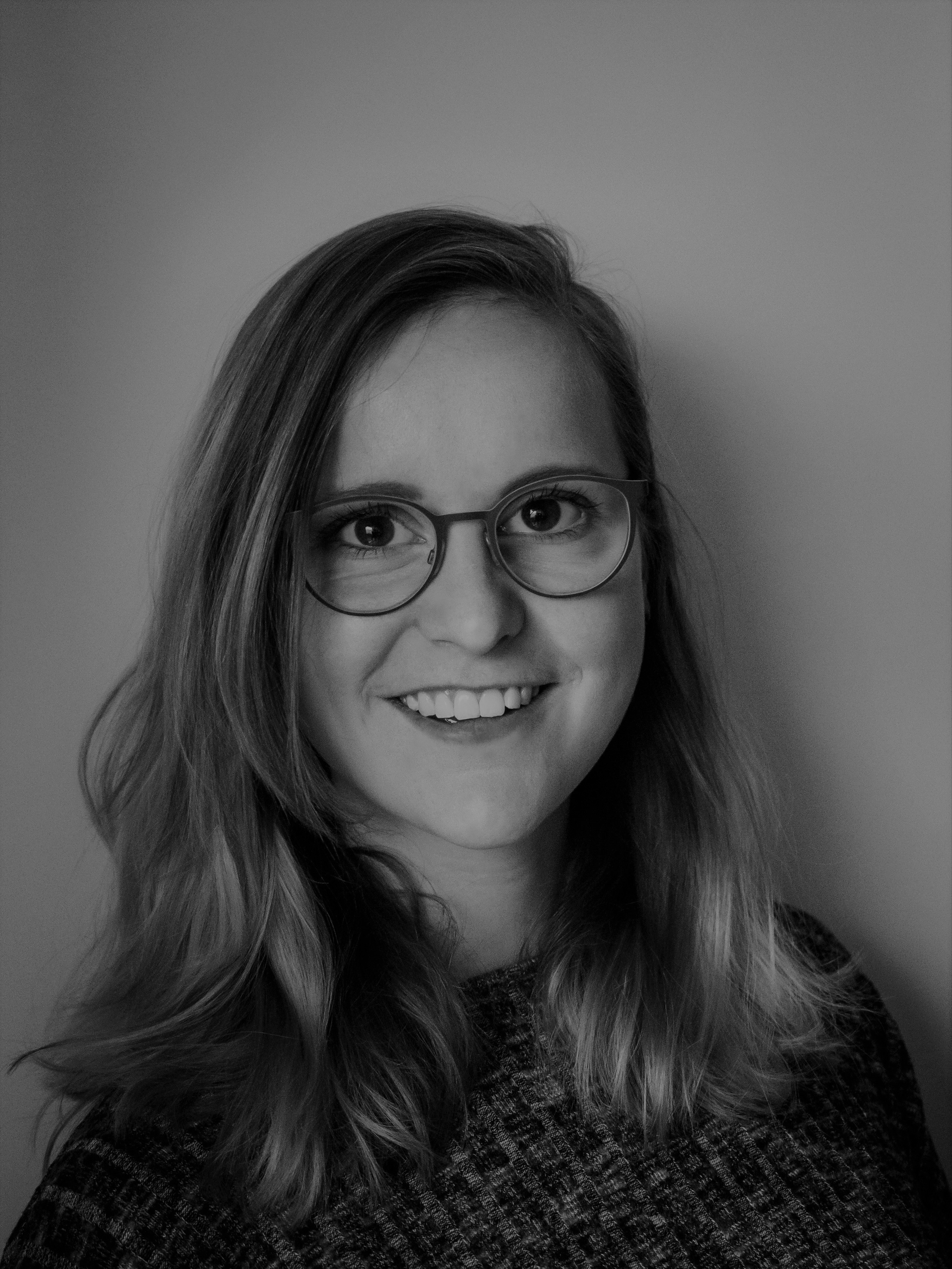}}]{Patricia Goldberg}
received her BSc degree in cognitive science from the University of Osnabrück and her MA degree in education science from the University of Freiburg, Germany. She did her PhD in Psychology at the Hector Research Institute of Education Sciences and Psychology at the University of Tübingen. In her research, she is focusing on attentional processes in teacher-learner interactions to improve teacher training and classroom instruction.
\end{IEEEbiography}

\begin{IEEEbiography}[{\includegraphics[width=1in,height=1.25in,clip,keepaspectratio]{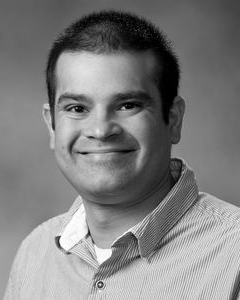}}]{Sidney D'Mello}
Sidney D'Mello (PhD in Computer Science) is an Associate Professor in the Institute of Cognitive Science and Department of Computer Science at the University of Colorado Boulder. He is interested in the dynamic interplay between cognition and emotion while individuals and groups engage in complex real-world tasks. He applies insights gleaned from this basic research program to develop intelligent technologies that help people achieve to their fullest potential by coordinating what they think and feel with what they know and do. D'Mello has co-edited seven books and published almost 300 journal papers, book chapters, and conference proceedings. His work has been funded by numerous grants and he currently serves(d) as associate editor for Discourse Processes and PloS ONE. D'Mello is the Principal Investigator for the NSF National Institute for Student-Agent Teaming.
\end{IEEEbiography}

\begin{IEEEbiography}[{\includegraphics[width=1in,height=1.25in,clip,keepaspectratio]{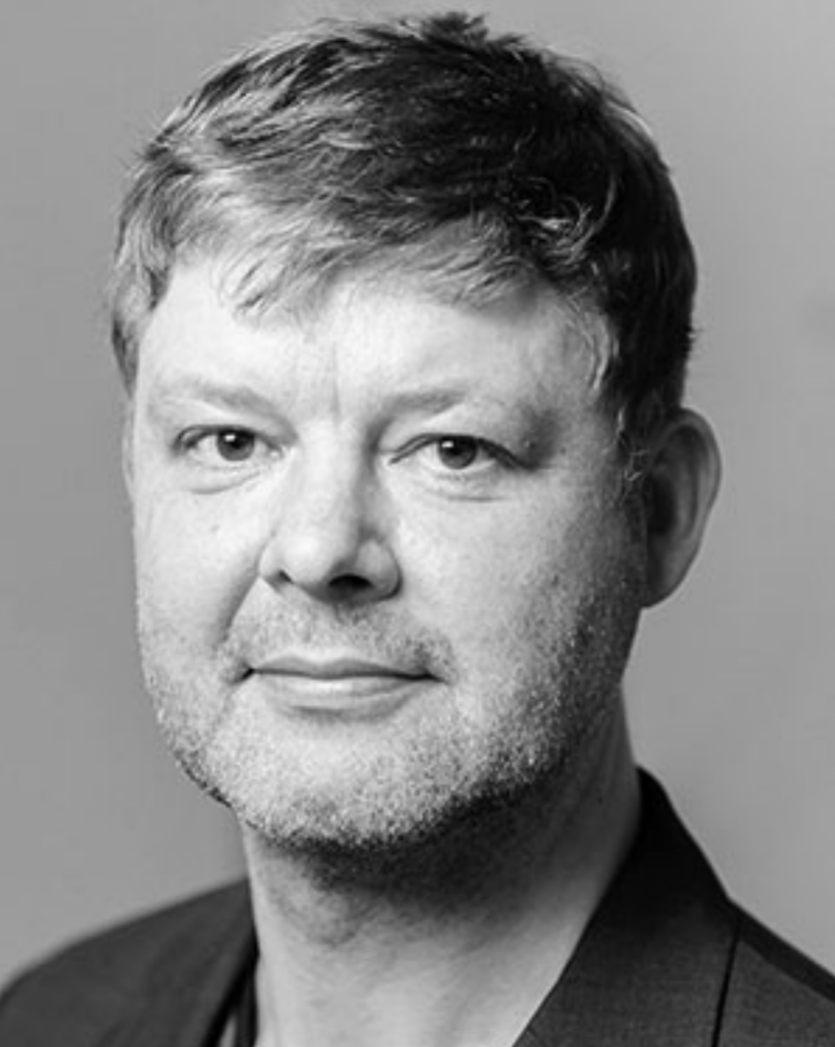}}]{Peter Gerjets}
received the diploma in psychology from the University of Gottingen, in 1991. From 1991 to 1995 he was a research associate with
the University of Gottingen where he received the PhD degree in 1994. Afterwards he has been working as assistant professor with the Saarland University in Saarbrucken where he finished his habilitation in 2002 before taking over his current position at the University of Tubingen. Since 2002 he has been working as principal research scientist at the Knowledge Media Research Center and beside as full professor for research on learning and instruction at the University of Tubingen. He was honoured with the Young Scientist Award of the German Cognitive Science Society in 1999 and served in the editorial boards of the Journal of Educational Psychology, the Educational
Research Review, the Computers in Human Behavior, and the Educational Technology, Research, and Development. His current research focuses on multimodal and embodied interaction with digital media as well as on learning from multimedia, hypermedia, and the Web. He is a member of DGPs, APS, and EARLI and served as coordinator of the EARLI Special Interest Group 6:Instructional Design.
\end{IEEEbiography}

\begin{IEEEbiography}[{\includegraphics[width=1in,height=1.25in,clip,keepaspectratio]{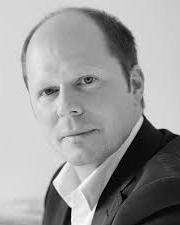}}]{Ulrich Trautwein}
is a Professor of Education Sciences and Executive Director of the Hector Research Institute of Education Sciences and Psychology at the University of Tübingen. He is also Co-Director of the LEAD Research Network. Prior to this, Professor Trautwein was a senior researcher at the Max Planck Institute for Human Development in Berlin. His area of research is empirical educational research, broadly defined.In particular, his research interests are directed primarily to the development of self-referent cognitions in the school context, school management and the influence of homework on school achievement. Trautwein has published a large number of articles on a number of topics, including the development of conscientiousness, expectancy-value beliefs, and academic effort, the effectiveness of homework assignments and completion, and the results of several randomized controlled field trials in school settings.
\end{IEEEbiography}

\vspace{-12cm}
\begin{IEEEbiography}[{\includegraphics[width=1in,height=1.25in,clip,keepaspectratio]{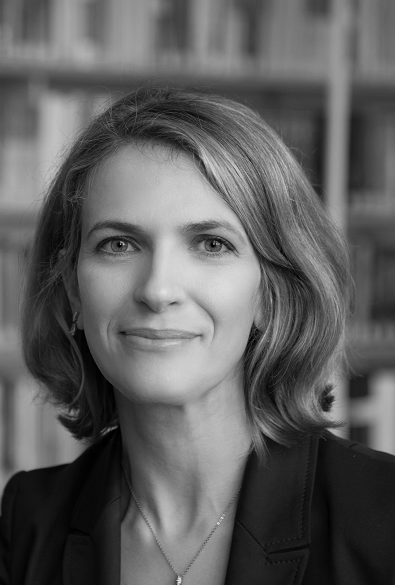}}]{Enkelejda Kasneci}
is a Professor of Computer Science at the University of Tübingen, Germany, where she leads the Human-Computer Interaction Lab. As a BOSCH scholar, she received her M.Sc. degree in Computer Science from the University of Stuttgart in 2007. In 2013, she received her PhD in Computer Science from the University of Tübingen. For her PhD research, she was awarded the research prize of the Federation Südwestmetall in 2014. From 2013 to 2015, she was a postdoctoral researcher and a Margarete-von-Wrangell Fellow at the University of Tübingen. Her research evolves around the application of machine learning for intelligent and perceptual human-computer interaction. She serves as academic editor for PlosOne and as a TPC member and reviewer for several major conferences and journals.
\end{IEEEbiography}


\end{document}